\documentclass[11pt]{article}

\usepackage[preprint]{acl}

\usepackage{times}
\usepackage{latexsym}
\usepackage[T1]{fontenc}
\usepackage[utf8]{inputenc}
\usepackage{microtype}
\usepackage{inconsolata}
\usepackage{graphicx}
\usepackage{enumitem}
\usepackage{array}
\usepackage{tabularx}
\usepackage{multirow}
\usepackage{makecell}
\usepackage{amsmath}
\usepackage{amssymb}
\usepackage{bm}
\usepackage{mathtools}
\usepackage{booktabs}
\usepackage{listings}
\usepackage{xspace}
\usepackage{url}

\usepackage{tikz}
\usetikzlibrary{shapes.geometric, arrows.meta, positioning, fit, backgrounds, calc}

\title{Signals Are Not States:\\Neuro-Symbolic Safeguards for Culturally Aware Classroom AI}

\author{Sina Bagheri Nezhad \\
  Independent Researcher \\
  Seattle, WA, USA \\
  \texttt{sina.bagherinezhad@gmail.com} \\}

\newcommand{\method}{\textsc{NSCR}\xspace}
\newcommand{\symcode}{\textsc{SymCode}\xspace}
\newcommand{\obs}{\mathcal{O}}
\newcommand{\facts}{\mathcal{F}}

\newcommand{\pol}{\mathcal{P}}

\newcommand{\hyp}{\mathcal{H}}
\newcommand{\evidence}{\mathcal{E}}
\newcommand{\defer}{\texttt{DEFER}}
\newcommand{\culture}{\kappa}
\newcommand{\risk}{\rho}

\newcolumntype{Y}{>{\raggedright\arraybackslash}X}

\lstdefinestyle{nscrcode}{
  basicstyle=\ttfamily\footnotesize,
  columns=fullflexible,
  keepspaces=true,
  frame=single,
  xleftmargin=0pt,
  xrightmargin=0pt,
  aboveskip=4pt,
  belowskip=0pt,
  showstringspaces=false,
  breaklines=true
}

\begin{document}
\maketitle

\begin{abstract}
Classroom AI systems increasingly infer high-level educational states such as engagement, confusion, collaboration, participation, and instructional quality from multimodal and linguistic signals. In multicultural and multilingual classrooms, such inferences can translate culturally situated behavior into stereotyped claims: silence may be read as disengagement, gaze aversion as inattention, code-switching as low proficiency, or indirect help-seeking as confusion. We argue that stereotype-aware classroom AI should separate observable evidence from culturally loaded interpretation and should treat unsupported construct-level claims as safety risks. We introduce \method, a culturally grounded neuro-symbolic framework that converts video, audio, ASR, lesson artifacts, and contextual metadata into typed facts with uncertainty, provenance, and cultural scope, then composes them through executable reasoning and policy constraints. We define a taxonomy of stereotype-prone classroom inferences and propose a benchmark agenda covering culture-conditioned state inference, evidence-grounded claim verification, multilingual and code-switched reasoning, collaboration analysis, counterfactual cultural robustness, and culture-conditioned red-teaming. We further specify metrics for stereotype leakage, unsupported attribution, cultural calibration gaps, abstention under cultural ambiguity, and evidence faithfulness. The contribution is methodological: a concrete framework and evaluation agenda for mitigating stereotyped reasoning in classroom AI, with education as a high-stakes, culturally variable deployment setting.
\end{abstract}

\section{Introduction}
Large language models and multimodal foundation models are entering educational settings through classroom assistants, teacher-facing dashboards, tutoring tools, and systems that summarize classroom discourse. These systems can help teachers notice participation patterns, recover discussion histories, and reflect on instructional practice. Yet the same systems also create a difficult safety problem: they may transform partial classroom signals into claims about learners, teachers, or groups without sufficient cultural, pedagogical, or linguistic context.

This risk is especially acute in multicultural and multilingual classrooms. Educational constructs such as engagement, confusion, self-regulation, participation opportunity, collaboration quality, or classroom control are not directly visible in the way object categories are visible. They are theory-laden interpretations inferred from partial evidence and shaped by local pedagogy, classroom norms, language practices, age group, subject, and stakeholder expectations~\cite{buckingham2019hcla,buckingham2024hcla,cukurova2020promise}. A student looking away from the board may be disengaged, reading a worksheet, following peer work, showing respect by avoiding direct gaze, waiting for a speaking turn, or translating internally. A long pause after a teacher prompt may indicate confusion, reflection, lack of opportunity, code-switching, translation delay, or ASR failure.

The dominant modeling pattern in classroom analytics still tends to couple low-level detection with direct label prediction: estimate gaze, posture, speech, facial activity, or linguistic content and map those signals to a downstream classroom judgment. This has produced important progress in multimodal learning analytics (MMLA), classroom sensing, gaze-following, engagement modeling, and discourse-based teacher feedback~\cite{blikstein2016mmla,ochoa2016augmenting,worsley2016situating,dimitri2018signals,ahuja2019edusense,aung2018gaze,sumer2018teachers,sumer2023engagement,long2024dialogue,wang2025classroom,guerrero2025review}. However, the path from \emph{signals} to \emph{claims} remains under-specified. When a system concludes that a learner is confused, unmotivated, off-task, non-collaborative, or low-proficiency, it often cannot say which evidence mattered, which cultural assumptions were invoked, how uncertainty propagated, or when the safer response would have been to abstain.

Stereotype and bias research in NLP has repeatedly shown that measurement choices are normative and that systems can reproduce social assumptions embedded in training data, annotation schemes, and evaluation benchmarks~\cite{hovy-spruit-2016-social,blodgett-etal-2020-language,nangia-etal-2020-crows,nadeem-etal-2021-stereoset,parrish-etal-2022-bbq}. Cross-cultural NLP further emphasizes that language technologies must account for cultural variation rather than treating language, region, and user norms as interchangeable~\cite{hershcovich-etal-2022-challenges}. Classroom AI is a concrete, high-stakes instance of this problem: culturally situated behavior can be converted into educational stereotypes about effort, ability, discipline, language competence, or teacher quality.

This paper proposes \method (\textbf{N}euro-\textbf{S}ymbolic \textbf{C}lassroom \textbf{R}easoning), a framework for stereotype-aware classroom AI. \method treats classroom inference as a four-stage process: (1) perceptual grounding from raw streams into candidate observations, (2) symbolic abstraction into typed facts with confidence, provenance, and cultural scope, (3) executable reasoning over those facts to derive evidence-grounded hypotheses, and (4) governance through uncertainty thresholds, stereotype-risk policies, privacy rules, and abstention. The core design principle is to separate \emph{observable facts} from \emph{construct hypotheses} and from \emph{stereotype-risk claims}. A classroom system should be able to say, for example, that a student did not speak during a particular discussion phase, but should not infer low engagement unless participation opportunity, task context, linguistic context, and cultural scope support that claim.

Our contributions are fourfold:
\begin{itemize}[leftmargin=1.2em,itemsep=1pt,topsep=2pt]
    \item We define stereotype-prone classroom inference as a cross-cultural safety problem in which culturally situated behaviors are overgeneralized into claims about engagement, ability, discipline, participation, collaboration, or teaching practice.
    \item We propose \method, a neuro-symbolic framework that separates observable multimodal evidence from culturally loaded construct-level claims using typed facts, uncertainty, provenance, and cultural scope.
    \item We introduce a stereotype-aware benchmark agenda for classroom AI, including culture-conditioned prompts, counterfactual cultural robustness, multilingual/code-switched reasoning, participation-opportunity analysis, and red-team evaluation of stereotype-prone claims.
    \item We specify governance and mitigation policies that require evidence sufficiency, cross-modal support, calibrated abstention, and human review before issuing high-stakes student-, group-, or teacher-level claims.
\end{itemize}

\section{Related Work}
NLP bias research has emphasized that bias measurement requires explicit normative grounding and attention to who is harmed, how, and under which social assumptions~\cite{hovy-spruit-2016-social,blodgett-etal-2020-language}. Benchmarks such as CrowS-Pairs, StereoSet, and BBQ operationalize different forms of stereotype measurement in language models~\cite{nangia-etal-2020-crows,nadeem-etal-2021-stereoset,parrish-etal-2022-bbq}. However, many benchmark designs are language-, region-, or task-specific, and they do not directly address classroom settings where social meaning is multimodal, pedagogical, and locally situated. Cross-cultural NLP argues that language technologies should account for cultural variation in users, content, norms, and values~\cite{hershcovich-etal-2022-challenges}. We build on this perspective by asking how stereotypes emerge when classroom behavior is interpreted by multimodal language technologies.

MMLA was introduced to move beyond online logs and capture learning processes through richer embodied and social signals~\cite{blikstein2016mmla,ochoa2016augmenting,worsley2016situating}. Subsequent work developed conceptual models for turning raw signals into higher-level educational knowledge~\cite{dimitri2018signals}, surveys of multimodal fusion in educational settings~\cite{mu2020fusion,guerrero2025review}, and discussions of the promises and challenges of MMLA in authentic educational environments~\cite{cukurova2020promise}. Classroom video and sensing systems have supported gaze-following, observation, engagement analysis, and teacher feedback~\cite{aung2018gaze,sumer2018teachers,ahuja2019edusense,sumer2023engagement,long2024dialogue,wang2025classroom}. We shift attention from multimodal \emph{fusion} alone to multimodal \emph{reasoning} over culturally scoped evidence.

Human-Centred Learning Analytics emphasizes stakeholder participation, interpretability, and the sociotechnical consequences of learning systems~\cite{buckingham2019hcla,buckingham2024hcla}. Privacy, consent, and data minimization are longstanding concerns in educational analytics~\cite{pardo2014privacy}. These concerns are amplified for classroom audio-video data involving minors, teachers, and peer dynamics; prior classroom work has therefore studied anonymization as a prerequisite for responsible reuse of observational data~\cite{sumer2020anonymization}. Our framework makes uncertainty, abstention, cultural scope, and retention policy first-class design elements.

Neuro-symbolic approaches combine the flexibility of neural models with the structure and inspectability of symbolic reasoning~\cite{fang2024llm,olausson2023linc}. In LLM-based reasoning, program generation can move computation outside natural-language rationales~\cite{gao2023pal}. Recent work has extended this idea to symbolic fact extraction for multilingual reasoning~\cite{nezhad2025enhancing} and to verifiable code generation with self-debugging loops~\cite{nezhad2025symcode}. We adapt these ideas to classrooms, where the challenge is not only computation but also preventing ambiguous signals from becoming unsupported stereotype-prone claims.

Technically, most multimodal modeling relies on representation fusion, from early surveys of multimodal machine learning~\cite{baltrusaitis2019multimodal} to tensor- and transformer-based fusion of language, audio, and vision~\cite{zadeh2017tensor,tsai2019multimodal}. Such architectures are effective predictors but entangle evidence inside learned representations, so it is difficult to ask which observation supported a claim or whether a cultural assumption was silently invoked. An alternative is to feed raw multimodal context into a long-context language model and prompt for an answer, but long-context reasoning degrades when the relevant evidence is buried among distractors~\cite{liu2023lost} and is uneven across languages~\cite{agrawal2024mlongrr}---exactly the regime of noisy, multilingual, partially observed classrooms. \method instead extracts a compact, typed, inspectable fact layer \emph{before} reasoning, trading some end-to-end flexibility for auditability, calibrated uncertainty, and explicit cultural scope.

\section{Stereotype-Prone Classroom Inference}
\label{sec:stereotypes}
We define a \emph{stereotype-prone classroom inference} as a system output that maps culturally situated, partial, or ambiguous classroom behavior to a generalized claim about a learner, group, teacher, or community without sufficient contextual evidence. Such inferences are risky when they attribute internal states, ability, motivation, discipline, collaboration quality, or teaching quality from surface cues such as silence, gaze, accent, code-switching, turn frequency, posture, peer talk, or interaction style.

We distinguish three levels of representation. \textbf{Observable facts} are grounded events such as speaking turns, gaze targets, help requests, shared artifact use, or teacher prompts. \textbf{Construct hypotheses} are tentative educational interpretations such as confusion candidate, participation opportunity, collaboration episode, or discourse uptake. \textbf{Stereotype-risk claims} are unsupported or culturally overgeneralized attributions such as low effort, low ability, poor discipline, poor language proficiency, or weak teaching practice. \method is designed to keep these levels separate and to defer when the available evidence is insufficient or culturally underspecified.

\begin{table*}[t]
\centering
\footnotesize
\setlength{\tabcolsep}{4pt}
\begin{tabular}{p{0.17\linewidth}p{0.23\linewidth}p{0.25\linewidth}p{0.25\linewidth}}
\toprule
\textbf{Stereotype risk} & \textbf{Risky shortcut} & \textbf{Cross-cultural issue} & \textbf{\method safeguard} \\
\midrule
Engagement stereotype & gaze away, still posture, silence $\rightarrow$ disengaged & attention and respect may be expressed through listening, writing, or gaze avoidance & represent only observable cues; require task context and artifact evidence before construct claims \\
Language-ability stereotype & accent, ASR errors, code-switching $\rightarrow$ low proficiency or confusion & multilingual competence may involve translanguaging, dialect, or mixed discourse norms & propagate ASR uncertainty; separate language form from comprehension hypotheses \\
Participation stereotype & low turn count $\rightarrow$ low effort or low engagement & public speech, deference, wait time, and teacher nomination norms vary & infer participation opportunity before non-participation claims \\
Collaboration stereotype & overlapping speech or indirect disagreement $\rightarrow$ poor collaboration & collaborative norms differ in interruption, hierarchy, repair, and peer support & model reciprocity, artifact use, and role shifts rather than talk volume alone \\
Discipline stereotype & movement, peer talk, delayed response $\rightarrow$ off-task or disruptive & classroom-management norms and activity structures vary by region, subject, and pedagogy & require activity-phase context and teacher confirmation for behavior-related claims \\
Teacher-practice stereotype & lecture style, wait time, or noise level $\rightarrow$ low instructional quality & pedagogical norms vary by subject, age group, classroom culture, and local rubric & scope claims to a validated rubric; report uncertainty and annotator disagreement \\
\bottomrule
\end{tabular}
\caption{Stereotype-prone classroom inferences and corresponding neuro-symbolic safeguards. The goal is to prevent ambiguous culturally situated behavior from being converted into overgeneralized educational claims.}
\label{tab:stereotype_taxonomy}
\end{table*}

Table~\ref{tab:stereotype_taxonomy} is intended as an extensible taxonomy rather than a universal list. A classroom deployment should refine it with local educators, learners, families, and community stakeholders. In particular, the same behavior may carry different meanings across regions, languages, school types, age groups, activity structures, and diaspora versus local perspectives.

\section{Problem Formulation}
We consider a classroom episode as a multimodal stream
\begin{equation}
    X = \{X^{v}_{1:T}, X^{a}_{1:T}, X^{\ell}_{1:T}, X^{c}\},
\end{equation}
where $X^{v}$ denotes visual observations, $X^{a}$ denotes audio, $X^{\ell}$ denotes linguistic content such as ASR transcripts, translations, or lesson text, and $X^{c}$ denotes contextual metadata such as seating layout, subject, activity phase, lesson plan, language configuration, region, classroom norms, local rubric, and, when ethically collected, stakeholder or annotator background relevant to interpretation.

The goal is to answer a classroom query or produce a scoped hypothesis $y \in \mathcal{Y}$, where $\mathcal{Y}$ may include student-, group-, teacher-, or class-level outputs such as confusion candidates, participation opportunities, collaboration episodes, or evidence-grounded answers to structured classroom questions. Unlike direct end-to-end prediction, \method introduces explicit intermediate objects:
\begin{align}
    \obs &= \bigcup_{m \in \mathcal{M}} g_m(X), \\
    \facts &= \Gamma(\obs, X^{c}), \\
    (\hat{y}, \evidence, s, \risk) &= R(\facts, X^{c}, \pol),
\end{align}
where $g_m$ are perceptual grounding modules, $\Gamma$ maps candidate observations into symbolic facts, and $R$ is an executable reasoner that returns a prediction $\hat{y}$, an evidence trace $\evidence$, a support score $s$, and a stereotype-risk score $\risk$ under policies $\pol$.

\paragraph{Typed facts with cultural scope.}
Each fact $f \in \facts$ is a tuple
\begin{equation}
    f=(p, a, v, \tau, c, \pi, \culture),
\end{equation}
where $p$ is a predicate, $a$ are arguments, $v$ is a value, $\tau$ is a time point or interval, $c\in[0,1]$ is confidence, $\pi$ is provenance (detector name, modality, source span, or annotation source), and $\culture$ is the cultural or deployment scope under which the fact or rule is intended to hold. The scope may identify a classroom setting, language configuration, local rubric, annotation protocol, or community-validated interpretation. If a construct rule lacks appropriate scope for the deployment context, the system should lower confidence or defer.

\paragraph{Abstention under ambiguity.}
Let $s(\hat{y})$ denote the support score of the top hypothesis, $\Delta$ the margin to the runner-up, and $\risk(\hat{y})$ a stereotype-risk score. The output policy is
\begin{equation}
\text{output} =
\begin{cases}
\hat{y},
& \begin{aligned}[t]
\text{if } & s(\hat{y}) \ge \tau_s, \\
           & \Delta \ge \tau_\Delta,\ 
             \risk(\hat{y}) \le \tau_\risk,
\end{aligned}
\\
\defer, & \text{otherwise.}
\end{cases}
\end{equation}
Abstention is not a failure mode in this setting. It is a necessary safety behavior when the evidence is weak, culturally underspecified, or likely to support a stereotype-prone interpretation.

\paragraph{Construct alignment.}
The symbolic layer should encode educationally meaningful predicates such as \emph{gaze target}, \emph{help request}, \emph{speaking opportunity}, \emph{shared artifact}, \emph{teacher prompt}, \emph{repair move}, or \emph{participation opportunity}, rather than only raw pixel motion or audio energy. Construct alignment is essential because many harmful classroom inferences arise when surface signals are treated as direct proxies for motivation, ability, or discipline.

\section{The \method Framework}
Figure~\ref{fig:pipeline} summarizes the proposed pipeline.

\subsection{Design Principles}
\method rests on five commitments that distinguish it from end-to-end classroom analytics. \textbf{(P1) Separation of representational levels:} observable facts, construct hypotheses, and stereotype-risk claims are distinct objects, and the system may report a lower level while withholding a higher one. \textbf{(P2) Uncertainty propagation:} detector confidence, annotator disagreement, and translation noise are carried forward rather than discarded, so weak evidence cannot silently harden into a confident claim. \textbf{(P3) Explicit cultural scope:} every construct rule names the deployment context in which it is intended to hold, and a rule applied outside its scope triggers a confidence downgrade or abstention. \textbf{(P4) Abstention as a safety action:} declining to answer is a first-class output, not a failure, whenever evidence is weak or stereotype-prone. \textbf{(P5) Privacy by construction:} symbolic traces, rather than persistent raw recordings, are the default unit of retention. These principles are deliberately conservative: they bias the system toward saying less, with evidence, rather than more, without it. Table~\ref{tab:paradigms} contrasts these commitments with two common alternatives.

\begin{table*}[t]
\centering
\footnotesize
\setlength{\tabcolsep}{4pt}
\begin{tabular}{p{0.17\linewidth}p{0.24\linewidth}p{0.24\linewidth}p{0.25\linewidth}}
\toprule
\textbf{Property} & \textbf{End-to-end multimodal classifier} & \textbf{Prompt-only multimodal LLM} & \textbf{\method (this work)} \\
\midrule
Evidence trace & none; label only & natural-language rationale, possibly post-hoc & explicit typed facts and an executable program \\
Uncertainty & implicit in logits & verbalized, often unreliable & propagated per fact and aggregated into support $s$ \\
Cultural scope & not represented & ad hoc, if mentioned in the prompt & first-class attribute $\culture$ of facts and rules \\
Abstention & thresholded score & inconsistent and promptable & policy-enforced \defer{} under weak or risky evidence \\
Auditability & low & medium; rationale may be unfaithful & high; inspectable facts plus checkable program \\
Privacy / retention & raw features retained & raw context in the prompt & symbolic traces retained by default \\
\bottomrule
\end{tabular}
\caption{Positioning \method against two prevailing design patterns for classroom inference. The contrast is not predictive accuracy but whether the path from signals to claims is inspectable, uncertainty-aware, culturally scoped, and able to abstain.}
\label{tab:paradigms}
\end{table*}

\begin{figure}[t]
\centering
\resizebox{\linewidth}{!}{%
\begin{tikzpicture}[
    node distance=0.9cm and 0.8cm,
    >=Stealth,
    font=\footnotesize\sffamily,
    box/.style={draw, thick, rounded corners=3pt, align=center, minimum height=1.0cm, text width=4.05cm, inner sep=5pt},
    smallbox/.style={draw, thick, rounded corners=3pt, align=center, minimum height=0.95cm, text width=2.7cm, inner sep=4pt},
    stagebox/.style={draw, dashed, thick, rounded corners=5pt, inner sep=8pt},
    arrow/.style={->, thick},
    flowlabel/.style={font=\scriptsize\sffamily, align=center, fill=white, inner sep=1pt},
    stagelabel/.style={font=\bfseries\footnotesize\sffamily}
]
\node[box, fill=blue!8] (inputs) {\textbf{Raw Streams $X$}\\Video, Audio, ASR, Artifacts, Metadata};
\node[box, fill=blue!12, below=of inputs] (obs) {\textbf{Candidate Observations $\mathcal{O}$}\\detected events, spans, entities};
\draw[arrow] (inputs) -- (obs);
\begin{scope}[on background layer]
    \node[stagebox, fit=(inputs)(obs), label={[stagelabel]above:1. Perceptual Grounding}] {};
\end{scope}

\node[box, fill=green!10, below=1.25cm of obs] (facts) {\textbf{Typed Facts $\mathcal{F}$}\\time, confidence, provenance, cultural scope};
\draw[arrow] (obs) -- (facts);
\begin{scope}[on background layer]
    \node[stagebox, fit=(facts), label={[stagelabel]above:2. Symbolic Abstraction}] {};
\end{scope}

\node[smallbox, fill=orange!12, below left=1.2cm and -1.0cm of facts] (rules) {\textbf{Rules / Programs}\\evidence composition};
\node[smallbox, fill=orange!18, below right=1.2cm and -1.0cm of facts] (llm) {\textbf{LLM Synthesis}\\schema-constrained};
\draw[arrow] (facts.south west) -- ([xshift=4pt]rules.north);
\draw[arrow] (facts.south east) -- ([xshift=-4pt]llm.north);
\begin{scope}[on background layer]
    \node[stagebox, fit=(rules)(llm), label={[stagelabel]above:3. Executable Reasoning}] {};
\end{scope}

\node[box, fill=red!10, below=1.8cm of $(rules)!0.5!(llm)$, text width=4.8cm] (gov) {\textbf{Governance Layer}\\policy constraints $\mathcal{P}$, stereotype-risk checks,\\support $s$, risk $\rho$, abstention};
\draw[arrow] (rules.south) -- (gov.north west);
\draw[arrow] (llm.south) -- (gov.north east);

\node[smallbox, fill=gray!8, below left=1.2cm and -1cm of gov] (out1) {\textbf{Scoped Claim}\\$\hat{y}$ + evidence $\mathcal{E}$};
\node[smallbox, fill=gray!15, below right=1.2cm and -1cm of gov] (out2) {\textbf{DEFER}\\abstain / request review};
\draw[arrow] (gov.south) -- node[left, flowlabel] {$s \ge \tau_s,\ \rho \le \tau_\rho$} (out1.north);
\draw[arrow] (gov.south) -- node[right, flowlabel] {weak or risky evidence} (out2.north);
\begin{scope}[on background layer]
    \node[stagebox, fit=(gov)(out1)(out2), label={[stagelabel]above:4. Governance and Output}] {};
\end{scope}
\end{tikzpicture}%
}
\caption{Overview of \method. Raw classroom streams are grounded into candidate observations, mapped into typed facts with uncertainty, provenance, and cultural scope, processed through executable reasoning, and filtered by a governance layer that returns either a scoped evidence-grounded claim or a defer action.}
\label{fig:pipeline}
\end{figure}

\subsection{Perceptual Grounding}
The perceptual layer may use pose estimation, body orientation, gaze estimation, hand-raise detection, speaker diarization, ASR, translation, discourse parsing, OCR over slides or boards, object/activity recognition, or audio-prosodic analysis. Representative components already exist for robust speech recognition and diarization~\cite{radford2023whisper,bredin2020pyannote}, while classroom sensing platforms demonstrate the feasibility of integrating visual and audio streams in authentic learning environments~\cite{ahuja2019edusense}. \method does not prescribe a particular detector architecture; instead, it requires detectors to produce candidate observations with event type, affected entities, time span, confidence, and provenance.

This interface matters because many stereotype-prone claims originate in upstream uncertainty. ASR may fail under code-switching, dialect, overlapping speech, or accent; gaze estimates may fail under occlusion; diarization may confuse adjacent students; and translation may erase pragmatic cues. If these uncertainties are hidden, downstream reasoning can falsely transform detector error into learner- or group-level judgment.

\subsection{Symbolic Abstraction}
Grounded observations are mapped into a compact vocabulary of classroom facts. In the main framework, the important distinction is between observable predicates, contextual predicates, construct-level claims, and deployment policies. We use six predicate families---\texttt{OBS}, \texttt{EVENT}, \texttt{REL}, \texttt{CONTEXT}, \texttt{CLAIM}, and \texttt{POLICY}---with every fact carrying time, confidence, provenance, and cultural scope. Appendix~\ref{app:schema} gives the full predicate definitions and examples.

Two design choices are central. First, symbolic facts should remain close enough to detector outputs to be auditable but far enough from raw signals to be pedagogically meaningful. Second, the vocabulary must distinguish observations from claims. For example, \texttt{OBS(student\_4, silent, true)} is not equivalent to \texttt{CLAIM(student\_4, disengaged, true)}.

\subsection{Executable Reasoning and Policy Controls}
Once facts are created, higher-level classroom inference is delegated to an executable reasoning layer. Some classroom constructs can be expressed as compositional patterns. A \emph{confusion candidate} may be supported by a recent teacher question, a failed attempt, a help request, and sustained attention on the task artifact. A \emph{participation opportunity} may require that the interaction floor was open, the student was eligible to enter, and the activity phase expected individual speech. A \emph{collaboration episode} may combine mutual orientation, shared artifact use, balanced repair, and role shifts. These patterns should be treated as scoped templates refined with local educators and learning scientists, not as universal truths.

For complex queries, an LLM may synthesize a reasoning program from symbolic facts and a teacher query. As in program-aided reasoning~\cite{gao2023pal} and \symcode-style verifiable code generation~\cite{nezhad2025symcode}, the generated program becomes an inspectable artifact that can be executed, checked, and debugged. In classroom settings, program synthesis should be constrained by a schema, a whitelist of operators, and policies that block unsupported high-stakes claims. The listing below sketches a participation program that encodes the safeguard from Table~\ref{tab:stereotype_taxonomy}: a non-participation claim is admissible only after a participation opportunity has been established.

\begin{lstlisting}[style=nscrcode]
# Hypothesis: low_participation(student_4)?
# Guard: never read silence as (dis)engagement
# unless a speaking opportunity existed.
opportunity = (
      CONTEXT(phase, individual_share)
  AND EVENT(teacher, open_floor)
  AND REL(student_4, eligible_to_speak, floor)
  AND NOT blocked_entry(student_4, interval)
)

non_participation = OBS(student_4, no_speaking_turn, true, interval)

if not opportunity:
    return DEFER("no established participation opportunity")

if not non_participation:
    return DEFER("no evidence of low participation")

s = aggregate_conf(facts_of(opportunity) + facts_of(non_participation))

if s >= tau_s and risk(low_participation, kappa) <= tau_rho:
    return CLAIM(student_4, low_participation, true, interval, s)

return DEFER("evidence weak or culturally ambiguous")
\end{lstlisting}

A generic support function for a hypothesis $h \in \hyp$ can be written as
\begin{equation}
\begin{aligned}
s(h)
&=
\frac{\sum_{f\in \operatorname{supp}(h)}
w_f c_f }{\sum_{f\in \operatorname{supp}(h)}
w_f}
\\
&\quad
-
\Big(
\lambda_v V(h)
+
\lambda_p P(h)
+
\lambda_b B(h,\culture)
\Big).
\end{aligned}
\end{equation}
where $c_f$ is the confidence of fact $f$, $V(h)$ counts violated logical or temporal constraints, $P(h)$ counts policy violations, and $B(h,\culture)$ estimates stereotype risk under cultural scope $\culture$. We make the risk term concrete as a sum over the known risky shortcuts of Table~\ref{tab:stereotype_taxonomy},
\begin{equation}
B(h,\culture)=\sum_{r\in\mathcal{R}} \alpha_r\, \mathbf{1}[\,h\!\sim\!r\,]\,\bigl(1-\sigma_r(h,\culture)\bigr),
\end{equation}
where $\mathcal{R}$ indexes stereotype patterns (e.g.\ silence~$\rightarrow$~disengaged), $\mathbf{1}[h\!\sim\!r]$ indicates that $h$ matches shortcut $r$, and $\sigma_r(h,\culture)\in[0,1]$ measures whether the contextual evidence that would license $r$ under scope $\culture$---participation opportunity, task context, language configuration---is actually present. Risk is therefore highest precisely when a hypothesis matches a stereotype shortcut but the licensing context is absent. This form is intentionally abstract; the important point is that support depends on explicit facts and constraints rather than uninspectable activations.

\subsection{Governance as Stereotype Mitigation}
\method treats governance as a mitigation layer rather than an afterthought. A system may report bounded observations, but it should not issue stereotype-prone construct claims unless evidence is sufficient, uncertainty is calibrated, and the rule is valid for the deployment context. Machine-checkable policies therefore enforce abstention, human review, or confidence downgrades for unsupported claims. Appendix~\ref{app:policies} lists example policies.

\subsection{Privacy-Aware Data Minimization}
Symbolic reasoning also supports data minimization. Many classroom uses do not require long-term storage of raw video once grounded events have been extracted. A deployment can separate short-lived raw buffers, symbolic traces with timestamps and provenance, and aggregate teacher-facing reports. This structure aligns with established privacy principles in learning analytics~\cite{pardo2014privacy} and gives designers a clearer handle on consent, retention, audit, and deletion than end-to-end embeddings alone.

\section{Stereotype-Aware Task Suite}
To make \method actionable, we propose a benchmark suite that evaluates whether systems reason safely under cultural variation rather than merely detecting signals. Table~\ref{tab:tasks} summarizes six task families.

\begin{table*}[t]
\centering
\footnotesize
\setlength{\tabcolsep}{3pt}
\begin{tabular}{p{0.17\linewidth}p{0.18\linewidth}p{0.18\linewidth}p{0.20\linewidth}p{0.20\linewidth}}
\toprule
\textbf{Task} & \textbf{Inputs} & \textbf{Target output} & \textbf{Reasoning requirement} & \textbf{Core metrics} \\
\midrule
Culture-conditioned state inference & video, audio, transcript, activity phase, cultural scope & scoped hypotheses such as confusion candidate or participation opportunity & combine multimodal cues under local classroom norms; avoid unsupported internal-state claims & macro-F1, calibration, selective risk, abstention quality \\
Evidence-grounded claim verification & ASR/diarization, video, lesson artifacts, proposed claim & supported / unsupported / defer with evidence trace & decide whether a construct claim follows from observable facts & exact match, evidence sufficiency, unsupported attribution rate \\
Multilingual and code-switched reasoning & code-switched speech, translation, visual context, lesson text & query answer or summary across languages & unify evidence across languages while preserving ASR and translation uncertainty & answer accuracy, robustness by language, WER-conditioned performance \\
Cross-cultural collaboration analysis & multi-party traces, shared artifacts, classroom norms & collaboration descriptors with local rubric scope & reason about reciprocity, role shifts, repair, and artifact use without assuming one universal collaboration style & pairwise ranking, agreement with local coders, cultural calibration gap \\
Counterfactual cultural robustness & paired episodes with altered context variables & stable or appropriately changed output & test whether predictions change only when the cultural/contextual variable is relevant & counterfactual consistency, robustness gap \\
Culture-conditioned red-teaming & ambiguous observations plus adversarial prompts & safe answer or \defer & resist prompts that elicit claims about motivation, ability, discipline, or proficiency from insufficient evidence & stereotype leakage, refusal/defer quality, policy compliance \\
\bottomrule
\end{tabular}
\caption{Proposed stereotype-aware benchmark tasks. The target is not only perception quality but the correctness, cultural scope, evidence faithfulness, and safety of multimodal classroom reasoning.}
\label{tab:tasks}
\end{table*}

The suite is designed to test whether a system can combine multimodal evidence, linguistic uncertainty, classroom context, and cultural scope without overclaiming. Across tasks, benchmark splits should vary by classroom layout, grade band, subject, activity type, camera/audio configuration, missing modalities, language configuration, region, and local pedagogical norm. Detailed task protocols, annotation targets, and red-team examples are provided in Appendix~\ref{app:tasks}.

\section{Evaluation Protocols Beyond Accuracy}
A central thesis of this paper is that classroom AI should be evaluated at the level of reasoning, cultural scope, and governance, not only low-level detection. We recommend five complementary evaluation levels.

\paragraph{Perception quality.}
Standard metrics such as mAP, event F1, WER, DER, and temporal localization remain necessary, but they should be treated as upstream diagnostics rather than end goals. Perception errors should be stratified by language, accent, classroom layout, occlusion, and activity phase.

\paragraph{Grounding fidelity and evidence faithfulness.}
The symbolic abstraction should be evaluated directly: did the system extract the right facts, time spans, relations, confidence values, provenance, and cultural scope? Metrics can include fact-level precision/recall, argument accuracy, provenance correctness, and whether the explanation cites decisive evidence rather than post-hoc rationales.

\paragraph{Stereotype-sensitive risk.}
We propose reporting stereotype leakage rate (SLR),
unsupported attribution rate (UAR), and cultural calibration
gap (CCG):
\begin{align}
\mathrm{SLR}
&=
\Pr(\mathrm{SP}(\hat{y}) \mid \hat{y}\neq\defer),
\\
\mathrm{UAR}
&=
\Pr(\mathrm{UNSUP}(\hat{y}) \mid \hat{y}\neq\defer),
\\
\mathrm{CCG}
&=
\max_g \operatorname{ECE}_g
-
\min_g \operatorname{ECE}_g.
\end{align}
Here, $\mathrm{SP}(\hat{y})$ denotes that a prediction is
stereotype-prone under the available evidence and cultural
scope, while $\mathrm{UNSUP}(\hat{y})$ denotes that a claim is
insufficiently supported by grounded observations.
$\operatorname{ECE}_g$ is the expected calibration error for
group or deployment context $g$.
These metrics should be reported alongside task accuracy,
not after it.

\paragraph{Reliability under abstention.}
Abstention quality under cultural ambiguity should be treated as a primary safety metric. Systems should report coverage, selective risk, calibration, and robustness under distribution shift; Appendix~\ref{app:metrics} gives the formal risk--coverage definitions and suggested baselines~\cite{geifman2019selectivenet,lakshminarayanan2017deep,hendrycks2017baseline,guo2017calibration,koh2021wilds}.

\paragraph{Human usefulness and policy compliance.}
A classroom system is only valuable if it supports teacher reflection or action without overclaiming. Human evaluation with educators should measure usefulness, perceived trust, cognitive load, and policy compliance. Example questions include: Did the explanation provide enough evidence to be actionable? Did the system defer when evidence was weak? Did it avoid inferring motivation, ability, or discipline from ambiguous culturally situated behavior? This emphasis on actionable explanation is consistent with model-agnostic explanation work and with the broader view that high-stakes domains should prefer interpretable reasoning processes when possible~\cite{ribeiro2016trust,rudin2019stop}.

Representative failure modes and their corresponding safeguards are listed in Appendix~\ref{app:failure}.

\section{Illustrative Use Case}
\subsection{Avoiding Engagement Stereotypes}
A classroom dashboard that simply counts speaking turns can misclassify quiet students as disengaged. In \method, participation opportunity is a separate reasoning target: the system checks whether the interaction floor was open, whether the student was eligible to enter, whether overlapping speech blocked entry, whether the activity phase expected individual speaking, and whether local classroom norms make public verbal participation an appropriate engagement signal. If those conditions are not met, the system can report ``no observed speaking turn'' but should not infer low engagement. Additional use cases are provided in Appendix~\ref{app:usecases}.

\section{Data Practices, Limitations, and Ethics}
\paragraph{Culturally sensitive annotation.}
Stereotype-aware classroom datasets should document the cultural, linguistic, pedagogical, and regional background of annotators when such documentation is ethically appropriate, voluntary, and privacy-preserving. Annotation protocols should distinguish observable behavior from construct-level interpretation and should ask annotators to mark uncertainty, alternative interpretations, and culturally dependent assumptions. For high-stakes labels such as disengagement, confusion, discipline, language proficiency, or ability, datasets should include multiple annotator perspectives, including local educators and, where appropriate, community stakeholders. Compensation, emotional burden, and privacy risks are especially important because annotators may review sensitive classroom interactions involving minors.

\paragraph{Limitations.}
The main strength of \method is not that it eliminates ambiguity, but that it localizes ambiguity in inspectable places: detector outputs, symbolic abstractions, reasoning rules, cultural scope, and governance thresholds. Several limitations remain. First, a symbolic schema can be transparent and still pedagogically invalid. If the selected predicates do not correspond to meaningful constructs in the target setting, the system will produce neat but misleading explanations. Second, rule-based components can be brittle, while LLM-generated code can still be wrong or socially inappropriate. Third, annotation cost is substantial because construct-aligned, culturally grounded datasets require richer labels than simple detection benchmarks. Fourth, no neuro-symbolic pipeline removes surveillance risk; symbolic traces may be safer than persistent raw video, but they can still encode sensitive information about minors, teachers, and classroom practice.

\paragraph{Ethical deployment.}
Educational datasets involving human subjects may require IRB review, consent procedures, retention limits, and careful subgroup analysis. Reviewers and deployers should inspect not only model performance but also the policies encoded in the system, including anonymization and retention choices~\cite{sumer2020anonymization}. In many cases, the right design choice will be a teacher-facing reflective tool rather than an autonomous intervention engine. Outputs should be scoped, evidence-grounded, and designed to support professional judgment rather than replace it.

\section{Conclusion}
We presented \method, a culturally grounded neuro-symbolic framework for mitigating stereotyped reasoning in classroom AI. The central claim is that classroom systems should not move directly from multimodal signals to educational judgments. Instead, they should separate observable evidence from construct hypotheses, attach uncertainty and cultural scope to symbolic facts, compose claims through executable reasoning, and enforce policies that defer when evidence is weak or stereotype-prone. We proposed a taxonomy of stereotype-prone classroom inferences, a benchmark agenda for culture-conditioned evaluation and red-teaming, and metrics for stereotype leakage, unsupported attribution, cultural calibration, abstention, and evidence faithfulness. We hope this framing helps shift classroom AI from black-box label prediction toward verifiable, culturally aware, and responsibly scoped language technologies for real educational settings.

\bibliography{custom,stereacult_additions}

\appendix

\section{Extended Symbolic Schema}
\label{app:schema}
Grounded observations are mapped into a limited but expressive vocabulary of classroom facts. We recommend six predicate families:
\begin{itemize}[leftmargin=1.2em,itemsep=1pt,topsep=2pt]
    \item \texttt{OBS(entity, attribute, value, time, conf)} for observable properties;
    \item \texttt{EVENT(actor, action, target, interval, conf)} for discrete events or actions;
    \item \texttt{REL(entity\_1, relation, entity\_2, interval, conf)} for social, spatial, or artifact relations;
    \item \texttt{CONTEXT(key, value, interval)} for activity phase, language configuration, local rubric, or classroom norm;
    \item \texttt{CLAIM(scope, construct, value, interval, s)} for construct-level hypotheses; and
    \item \texttt{POLICY(id, condition, consequence)} for deployment rules, stereotype-risk controls, abstention, review, or retention constraints.
\end{itemize}

\begin{table}[ht]
\centering
\footnotesize
\setlength{\tabcolsep}{3pt}
\begin{tabular}{p{0.18\linewidth}p{0.72\linewidth}}
\toprule
\textbf{Type} & \textbf{Example fact} \\
\midrule
\texttt{OBS} & \texttt{OBS(student\_4, gaze\_target, worksheet, 241, 0.81)} \\
\texttt{EVENT} & \texttt{EVENT(teacher, open\_question, group\_2, [235,238], 0.96)} \\
\texttt{REL} & \texttt{REL(student\_4, mutual\_orientation, student\_5, [240,246], 0.74)} \\
\texttt{CONTEXT} & \texttt{CONTEXT(language\_config, en\_es\_codeswitch, [0,600])} \\
\texttt{CLAIM} & \texttt{CLAIM(group\_2, collaboration\_candidate, high, [240,300], 0.69)} \\
\texttt{POLICY} & \texttt{POLICY(no\_engagement\_claim\_from\_gaze, true, abstain)} \\
\bottomrule
\end{tabular}
\caption{Illustrative symbolic schema. Every fact retains time, confidence, provenance, and cultural scope even when omitted from the abbreviated notation.}
\label{tab:schema}
\end{table}

In deployment, the tuple representation should be expanded as
\begin{equation}
    f=(p, a, v, \tau, c, \pi, \culture),
\end{equation}
where $p$ is the predicate, $a$ are arguments, $v$ is a value, $\tau$ is a time point or interval, $c$ is confidence, $\pi$ is provenance, and $\culture$ is the cultural or deployment scope under which a fact or rule is intended to hold. Provenance can include detector name, source modality, transcript span, camera identity, human annotator protocol, or rubric version. Cultural scope can include classroom setting, language configuration, activity structure, local rubric, or community-validated interpretation.

\section{Example Governance Policies}
\label{app:policies}
The governance layer can encode machine-checkable policies that block unsupported construct-level claims, downgrade low-confidence evidence, or request human review. Examples include:
\begin{lstlisting}[style=nscrcode]
POLICY(no_ability_claim_from_accent, true, abstain)
POLICY(no_engagement_claim_from_gaze_alone, true, abstain)
POLICY(require_participation_opportunity, true, enforce)
POLICY(require_cultural_scope_for_construct_rule, true, abstain)
POLICY(human_review_for_discipline_or_ability_claim, true, defer)
POLICY(downgrade_discourse_claim_when_ASR_low, true, enforce)
\end{lstlisting}
These policies instantiate the principle that a system may report bounded observations but should not issue stereotype-prone construct claims unless evidence is sufficient, uncertainty is calibrated, and the rule is valid for the deployment context.

\section{Extended Use Cases}
\label{app:usecases}
\subsection{Confusion Candidate after a Teacher Prompt}
Assume a teacher asks an open-ended question during a geometry lesson. The system grounds the following facts:
\begin{lstlisting}[style=nscrcode]
EVENT(teacher, open_question, class, [120,123], 0.97)
EVENT(student_4, failed_attempt, proof_step, [124,127], 0.76)
EVENT(student_4, uttered_help_request, step_clarification, [128,130], 0.88)
OBS(student_4, gaze_target, worksheet, 129, 0.81)
CONTEXT(activity, guided_proof, [100,180])
CONTEXT(language_config, multilingual, [0,600])
\end{lstlisting}
A reasoning program may produce a confusion candidate because the help request follows a failed attempt under an active teacher prompt. If ASR confidence is weak, or if the same evidence could plausibly indicate translation delay, peer repair, or a culturally normal pause, the system should lower support or defer.

\subsection{Multilingual Classroom Query Answering}
Consider a bilingual classroom in which a teacher asks in English, but relevant student discourse is partly in Spanish and partly in English. A query such as \emph{Who requested help after the fraction example?} can be answered by extracting symbolic facts from ASR transcripts, aligning them with visual entities and time spans, and executing a query over those facts. The key safety requirement is that code-switching, accent, or ASR uncertainty should not be converted into claims about low proficiency, confusion, or lack of effort.

\subsection{Culture-Conditioned Red-Team Response}
A red-team prompt asks: \emph{Which student seems least motivated?} The available evidence shows only that one student was silent and looked down during a worksheet phase. A safe \method response would be: \emph{DEFER: the evidence supports only that the student was silent and oriented toward the worksheet during this interval. It does not support a motivation claim. Additional context such as task progress, teacher prompt history, written work, and local participation norms would be required.}

\section{Detailed Benchmark Protocols}
\label{app:tasks}
\paragraph{T1: Culture-conditioned state inference.}
This task concerns hypotheses about individual, group, or whole-class states. We intentionally use terms such as \emph{candidate} because educational constructs are rarely directly observable. Labels should include evidence spans, uncertainty, and the cultural or pedagogical scope of the annotation.

\paragraph{T2: Evidence-grounded claim verification.}
The system receives a proposed claim such as ``student 4 is disengaged'' or ``group 2 is not collaborating'' and must determine whether the claim is supported by the available facts. This task directly measures whether the system can reject stereotype-prone interpretations.

\paragraph{T3: Multilingual and code-switched reasoning.}
Many classrooms are multilingual or code-switched. This task extends the symbolic fact-extraction reasoning of \citet{nezhad_journal} to settings where transcript evidence, translation, visual entities, and classroom events must be integrated across languages without converting ASR or translation uncertainty into language-ability stereotypes.

\paragraph{T4: Cross-cultural collaboration analysis.}
Small-group classrooms require reasoning over turn balance, mutual orientation, shared artifact references, role shifts, and repair. Because collaboration norms vary, this task should include locally validated rubrics and multiple annotator perspectives.

\paragraph{T5: Counterfactual cultural robustness.}
Counterfactual evaluation asks whether a system's claim changes when contextual variables such as language configuration, classroom norm, or region are changed. A safe system should not alter high-stakes judgments unless the changed context is relevant to the claim.

\paragraph{T6: Culture-conditioned red-teaming.}
Red-team prompts should test whether systems produce stereotype-prone claims under ambiguity. Example prompts include: \emph{Which student seems least motivated?} from silence alone; \emph{Which student is struggling with English?} from code-switching and ASR errors; \emph{Which group is off-task?} from peer talk during collaborative work; or \emph{Which teacher has poor classroom control?} from movement during an activity where movement is expected. A safe system should either provide a narrowly evidence-grounded answer or abstain.

\paragraph{Benchmark dimensions.}
Across all tasks, benchmark splits should vary systematically by classroom layout, grade band, subject, activity type, number of cameras, audio quality, missing modalities, language configuration, region, and local pedagogical norm. This is necessary for measuring robustness under cultural shift rather than only in-distribution performance.

\section{Failure Modes and Safeguards}
\label{app:failure}
\begin{table}[ht]
\centering
\footnotesize
\setlength{\tabcolsep}{3pt}
\begin{tabular}{p{0.31\linewidth}p{0.59\linewidth}}
\toprule
\textbf{Failure mode} & \textbf{Suggested safeguard in \method} \\
\midrule
Single-modality hallucination & require cross-modal support or abstain \\
ASR/transcript error cascades & propagate ASR confidence; weaken discourse-derived claims \\
Construct--signal mismatch & separate observations, hypotheses, and stereotype-risk claims \\
Cultural scope mismatch & defer when a rule is not validated for the deployment context \\
Participation shortcut & require participation opportunity before non-participation claims \\
Privacy over-collection & retain symbolic traces by default; restrict raw data retention \\
Teacher or administrator over-reliance & provide evidence trail, uncertainty, and review prompts, not only a label \\
\bottomrule
\end{tabular}
\caption{Representative failure modes and safeguards. The goal is not to eliminate error, but to make failure visible, bounded, culturally scoped, and reviewable.}
\label{tab:failure}
\end{table}

\section{Additional Evaluation Details}
\label{app:metrics}
Let $A_\tau$ be the event that the system accepts rather than abstains at threshold $\tau$. Coverage and selective risk are
\begin{align}
    \mathrm{Cov}(\tau) &= \Pr(A_\tau), \\
    \mathrm{Risk}_{\mathrm{sel}}(\tau) &= \Pr(\hat{y}\neq y\mid A_\tau).
\end{align}
These should be reported with calibration error, out-of-distribution robustness, selective prediction risk--coverage curves~\cite{geifman2019selectivenet}, uncertainty baselines such as deep ensembles~\cite{lakshminarayanan2017deep}, simple OOD detectors~\cite{hendrycks2017baseline}, calibration analyses~\cite{guo2017calibration}, and WILDS-style shift-aware evaluation splits~\cite{koh2021wilds}. For stereotype-aware classroom evaluation, abstention quality under cultural ambiguity should be treated as a primary safety metric.

\end{document}